%% file: main.tex
\documentclass[conference]{IEEEtran}

\usepackage{amsmath}

\usepackage[T1]{fontenc}

\usepackage{graphicx}

\usepackage{hyperref}

\usepackage{enumitem}

\usepackage{multirow}

\usepackage[numbers]{natbib}

\usepackage{subcaption}
\usepackage[export]{adjustbox}
\makeatletter
\renewcommand{\subsection}{%
  \@startsection{subsection}{2}{\z@}%
                {-3.25ex\@plus -1ex \@minus -.2ex}%
                {1.5ex \@plus .2ex}%
                {\normalfont\centering}}
\makeatother

\usepackage{tikz}
\usetikzlibrary{arrows.meta, positioning, shapes.geometric}

\tikzset{
  block/.style={
    rectangle,
    draw,
    minimum width=2cm,
    minimum height=1cm,
    align=center
  },
  line/.style={
    draw,
    -Latex
  }
}

\title{Exposing Image Classifier Shortcuts with \\ Counterfactual Frequency (CoF) Tables}

\author{%
    James Hinns\textsuperscript{1} and David Martens\textsuperscript{1}
}

\date{\textsuperscript{1}University of Antwerp}

\begin{document}

\maketitle

\begin{abstract}
  The rise of deep learning in image classification has brought unprecedented accuracy but 
  also highlighted a key issue: the use of `shortcuts' by models. Such shortcuts are 
  easy-to-learn patterns from the training data that fail to generalise to new data. 
  Examples include the use of a copyright watermark to recognise horses, snowy background to 
  recognise huskies, or ink markings to detect malignant skin lesions. The explainable AI (XAI) community has suggested using instance-level explanations 
  to detect shortcuts without external data, but this requires the examination of many 
  explanations to confirm the presence of such shortcuts, making it a labour-intensive process.
  To address these challenges, we introduce Counterfactual Frequency (CoF) tables, a novel 
  approach that aggregates instance-based explanations into global insights, and exposes shortcuts. 
  The aggregation implies the need for some semantic concepts to be used in the explanations, 
  which we solve by labelling the segments of an image. We demonstrate the utility of CoF tables 
  across several datasets, revealing the shortcuts learned from them.

\end{abstract}

\input{Sections/1-introduction}

\input{Sections/2-related_work}

\input{Sections/3-methodology}

\input{Sections/4-results}

\input{Sections/5-end}

\bibliographystyle{plainnat}
\bibliography{ref}

\input{Sections/appendix}

\end{document}

%% file: Sections/1-introduction.tex
\section{\large Introduction}

A problem that has long plagued Deep Learning (DL) is a lack of generalisation~\cite{geirhos2020shortcut,lapuschkin2019unmasking,me,d2022underspecification}.
Often models perform well in testing, but when put into deployment fail to recreate these results.
One reason is a model's tendency to exploit non-causal correlations, or `shortcuts', in the training data. 
These unwanted decision rules lead to poor performance on unseen data.
A parallel can be drawn from the tendency of models to learn shortcuts to the `principle of least effort' in linguistics~\cite{geirhos2020shortcut}. 
This principle suggests that language users and, by analogy, machine learning models, prefer the path of minimum effort. 
In the context of DL, this means that models may optimise towards simpler, albeit less generalisable, rules over those that are more complex so long as they still perform well on existing evaluations.

\begin{table}[h]
    \centering
    \resizebox{0.6\linewidth}{!}{%
    \begin{tabular}{ll}
        \hline
        \textbf{Segment Name} & \textbf{Frequency} \\ \hline
        watermark & 50\% \\ 
        grass & 25\% \\
        zebra & 25\% \\ \hline
    \end{tabular}%
    }
    \caption{Hypothetical CoF table for the watermark zebra example, where 50\% of the produced counterfactuals are based on the watermark, and 25\% are based on grass and zebra segments. The watermark serves as a shortcut for classifying horses, indicating its removal frequently results in a counterfactual.}
    \label{table:watermark_zebra_example}
\end{table}

Shortcuts have been demonstrated in a variety of applications within Image Classification. 
Notably, models have exploited watermarks as misleading cues in datasets like ImageNet \cite{Li_2023_CVPR_Whac_A_Mole} and Pascal VOC~\cite{lapuschkin2019unmasking}. 
Other instances include the reliance on surgical skin markings for melanoma classification \cite{winkler2019association} and the emphasis on body position for detecting COVID-19 in chest radiographs \cite{degrave2021ai}.

Often these shortcuts go unnoticed within standard model development cycles as they manifest in both training and testing sets. 
While testing the model with additional data from a broader range of sources can aid in identifying these shortcuts, acquiring sufficiently diverse data poses its own challenge.
In addition to the common challenges of data collection, many shortcuts naturally occur.
For example, since boats are frequently photographed on water, this scenario can introduce a background shortcut based on the water.


To combat this, it has been suggested to use instance-based explanation methods to identify unwanted rules learnt by models without the need for further data~\cite{lapuschkin2019unmasking,degrave2021ai,winkler2019association}.
Specifically within Image Classification, these explanations tend to be in the form of a saliency map, showing which pixels in the image the model is relying upon most for its decision. 
However identifying unwanted rules through individual instance-based explanations involves experts manually reviewing many explanations, to find those that exhibit unwanted decision rules.
This issue is exacerbated by the fact that some shortcuts only exist in a small portion of images, making the problem even more time consuming to diagnose (if diagnosed at all).
Compounding this issue is the nature of saliency maps, which, while highlighting the positions deemed important by the model, fail to represent other factors such as the semantics or texture of segments, along with pixel color and intensity. We argue that in many cases, these aspects are crucial for understanding the deeper reasoning behind a model's focus on certain segments of an image, as they can reveal more about the model's learned associations and decision-making processes.

In this study, we propose a solution to these problems, through the introduction of Counterfactual Frequency (CoF) tables, 
a novel analytical tool designed to aggregate and quantify the impact of semantically similar segments on the 
decision-making process of image classification models.  
These segments are identified by a segmentation model across a dataset of images 
$I = \{i_1, i_2, \ldots, i_n\}$.
For each image $i_x$ within $I$, a segmentation model delineates a set of $m$ segments 
$S_x = \{S_x^1,S_x^2, \ldots, S_x^m\}$ each assigned a name based on its semantic characteristics. 
CoF tables are based on the frequency with which these segments, identified by their common names,
cause changes in the predicted class when edited, resulting in a counterfactual.

By cataloguing how often specific segments, identified through their names, contribute to 
counterfactual explanations, CoF tables provide a systematic method for identifying the 
most influential factors within images that affect a model’s decision-making process.
This process highlights potential shortcuts in a model's classifications, facilitating a deeper 
understanding of the model's reliance on certain visual features. Through this aggregation, 
CoF tables aid in uncovering broader patterns of model behaviour, thereby 
contributing to the development of more robust and interpretable image classifiers.

The aggregation organised by CoF tables not only exposes the specific segments that are prone to 
causing classification changes when altered but also underscores the collective influence of these 
segments across the dataset.

In order to create CoF tables, we require explanations with semantics, not just small sets of pixels in proximity, which are not common in image explanations. Thus, we additionally present \textbf{S}emantic \textbf{C}ounterfactuals for \textbf{A}ccurate \textbf{P}icture (SCAP) explanations; a framework to produce counterfactual explanations with semantic meaning. SCAP explanations provide labels for the segments in an image that cause counterfactuals, which we can then use to create CoF tables.

\noindent Our main contributions are summarised as follows:
\begin{itemize}[noitemsep,topsep=0pt]
    \item \textbf{S}emantic \textbf{C}ounterfactuals for \textbf{A}ccurate \textbf{P}icture (SCAP) Explanations: A model-agnostic framework to produce counterfactual explanations for image classification given a way to segment images, and an edit function to change the segments provided.
    \item \textbf{Co}unterfactual \textbf{F}requency (CoF) tables: A method to aggregate local counterfactual explanations into global insights.
    \item Demonstration of SCAP explanations and CoF tables on various datasets, showcasing learned background shortcuts from foundational models trained on ImageNet.
\end{itemize}

%% file: Sections/2-related_work.tex
\section{\large Aggregating Local Explanations}



A number of explanation techniques have been proposed that can aggregate local explanations for tabular data.
These techniques often employ feature attribution (or importance) explanations, which can 
then be averaged to provide a simple global explanation~\cite{van2019global,lundberg2020local,me} of the model on the given data.
Notably, these aggregations can be visualised using the same techniques that SHAP~\cite{lundberg2017unified} employs for illustrating local explanations.




For image explanations, one of the most common categories is pixel attribution methods, also known as saliency maps.
Pixel attribution methods follow the same principle as feature attribution methods; both explain individual predictions by attributing decisions to input features, which, in the case of image classification, are typically individual pixels.
Two subdivisions of pixel attribution methods are occlusion-based methods and gradient-based methods~\cite{molnar2019}.
Occlusion-based methods systematically occlude different parts of the input image and observe the change in the output. Examples include SHAP~\cite{lundberg2017unified}, LIME~\cite{ribeiro2016should}, and SEDC-T~\cite{vermeire2022explainable} on which our SCAP explanations are based on.
SHAP can explain the attribution by segments, instead of pixels, as can SEDC-T, forming counterfactuals from segments within images, for easier interpretation of the produced saliency maps.
Gradient-based methods, such as Vanilla Gradient~\cite{simonyan2013deep} and Grad-CAM~\cite{selvaraju2017grad}, on the other hand, calculate the gradient of a prediction with respect to input features. These methods tend to be much less computationally expensive than occlusion-based methods, contributing to their greater popularity for image explanations.
Both these categories aim to explain \textit{which} pixels contribute most to a model's decision, but they do not reveal \textit{how} these pixels influence the decision.

However, pixel attribution methods face several significant challenges. 
Like most explanation methods, it is often extremely difficult to evaluate their accuracy.
They have been shown to produce different explanations for the same prediction and gradient, raising questions about their reliability~\cite{kindermans2019reliability}.
Furthermore, they may provide misleading views of a model's decision-making process~\cite{atrey2019exploratory}.
These methods can also be fragile, with small perturbations to an image leading to changes in explanations without corresponding changes in the model's predictions~\cite{ghorbani2019interpretation}.
There is even a discussion around some methods producing explanations that aren't related to the model or data~\cite{adebayo2018sanity}.

Although aggregation by averaging, as used in tabular data, is possible, providing only a single saliency map for many images can pose interpretation problems. 
Consider, once more, the example of a watermark shortcut in the binary classification of horses and zebras. 
Averaging the watermark positions (which the saliency map should focus on in the case of an identified shortcut) across many images would obscure this behaviour, as it would average the position of all watermarks, which could exist at any point within the image. 
This can be problematic because, in many shortcuts, the type of object in the image is the shortcut rather than its position~\cite{nauta2021uncovering}.




To more effectively address this challenge, SpRAy~\cite{lapuschkin2019unmasking} proposes combining multiple saliency maps through clustering instead of averaging. 
SpRAy identifies similar maps using spectral clustering and combines them to generate a number of aggregated explanations. 
Whilst this method demonstrates potential for shortcut identification, it relies on the presence of shortcut-inducing features in consistent positions. 
As shown in Figure~\ref{fig:SprayVis}, SpRAy can prioritise the position over the shape of objects used as shortcuts. This limitation arises from the inherent focus of pixel attribution methods on the positioning of significant pixels, as previously discussed.

%% file: Sections/3-methodology.tex
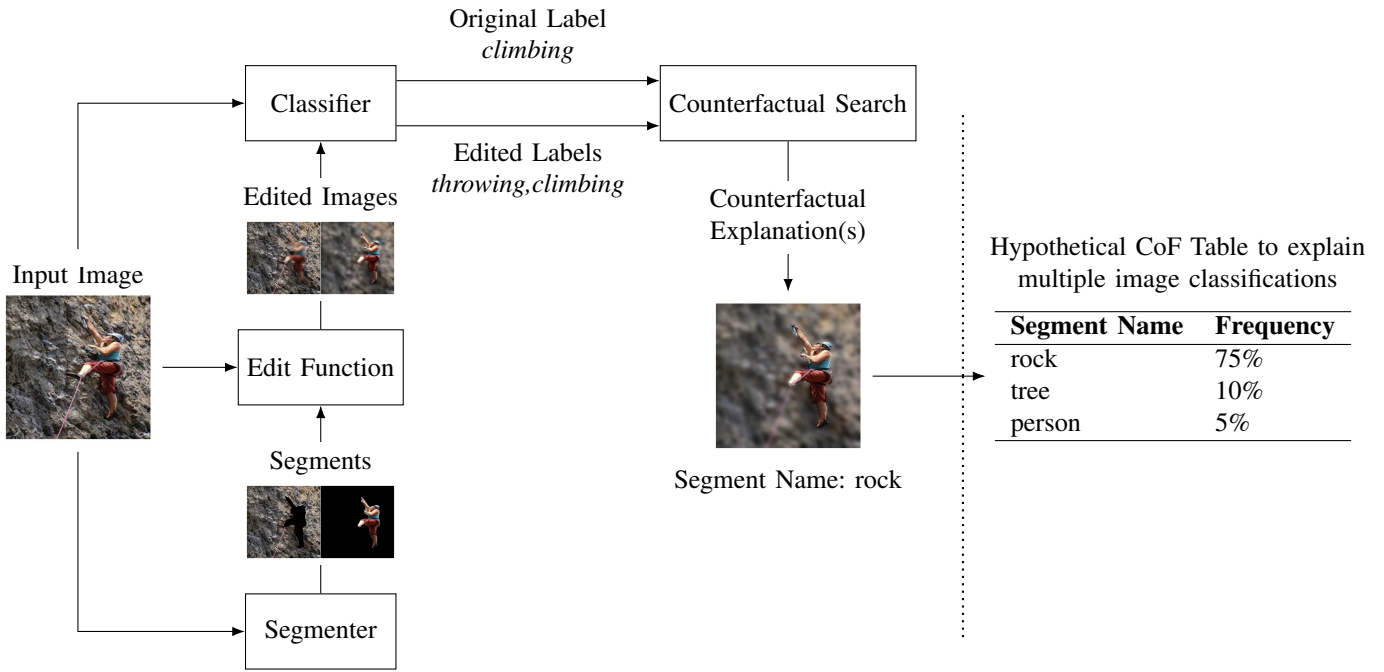
\begin{figure*}[hbt!]
    \centering
    \begin{tikzpicture}
        \node (image) {\includegraphics[width=2cm,height=2cm]{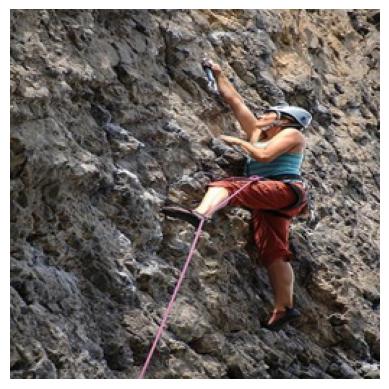}};
        \node [above=-0.2cm of image, align=center] {Input Image};
        
        \node [block, right=1cm of image] (edit) {Edit Function};
        \node [block, above=2.5cm of edit] (classifier) {Classifier}; 
        \node [block, below=2.5cm of edit] (segmenter) {Segmenter}; 
        \node [block, right=3.5cm of classifier] (counterfact) {Counterfactual Search};
        
        \node [below=2cm of counterfact] (explanations) {\includegraphics[width=2cm,height=2cm]{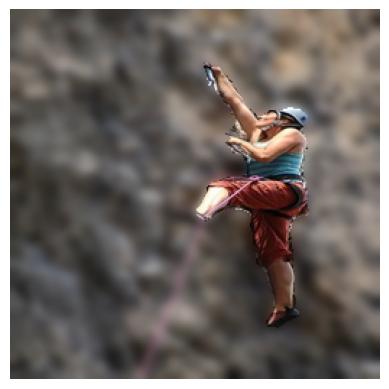}};
        
        \node [fill=white, above=0.2cm of image] (imTop) {};
        
        \draw [line] (image) |- (segmenter);
        \draw [line] (imTop) |- (classifier);
        \draw [line] (image) -- (edit);
        
        \draw [line] (segmenter) -- node[midway, fill=white, inner sep=2pt, yshift=-0.3cm] {\includegraphics[width=2cm,height=1cm]{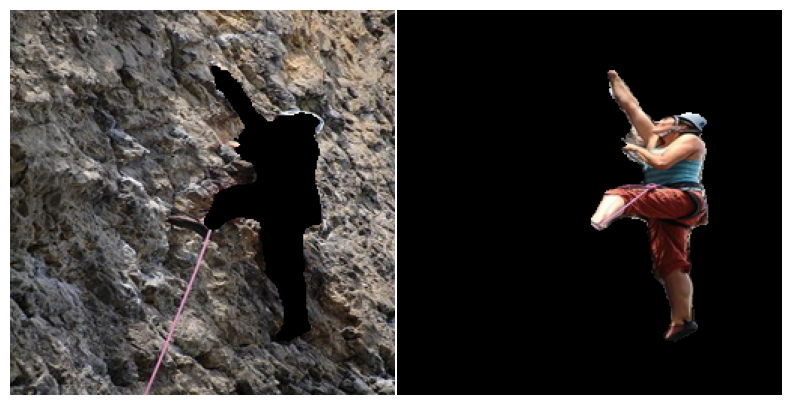}} node[midway, fill=white, above=0.2cm] {Segments} (edit);
        
        \draw [line] (edit) -- node[midway, fill=white, inner sep=2pt, yshift=-0.3cm] {\includegraphics[width=2cm,height=1cm]{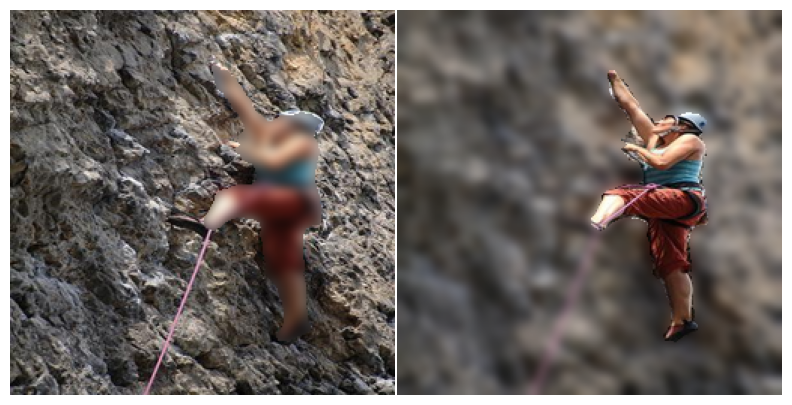}} node[midway, fill=white, above=0.2cm] {Edited Images} (classifier);
        
        \draw [line] ([yshift=3mm]classifier.east) -- node[midway, fill=white, above=0.1cm, align=center] 
        {Original Label\\ \textit{climbing}} ([yshift=3mm]counterfact.west);
        \draw [line] ([yshift=-3mm]classifier.east) -- node[midway, fill=white, below=0.1cm, align=center] 
        {Edited Labels\\ \textit{throwing,climbing}} ([yshift=-3mm]counterfact.west);
        
        \draw [line] (counterfact) -- (explanations);
        
        \node [below=0cm of explanations, align=center] {Segment Name: rock};
        \node [above=0.5cm of explanations, align=center, fill=white] {Counterfactual \\ Explanation(s)};
        
        \draw [dotted,thick] ([xshift=1.2cm]explanations.east) -- ++(0,3.5cm);
        \draw [dotted,thick] ([xshift=1.2cm]explanations.east) -- ++(0,-3.5cm);
        
        \node [right=1.5cm of explanations, align=center] (cof) {
            \begin{tabular}{ll}
            \hline
            \textbf{Segment Name} & \textbf{Frequency} \\ \hline
            rock          & 75\%           \\
            tree          & 10\%             \\
            person        & 5\%             \\ \hline
            \end{tabular}
        };
        \draw [line] (explanations.east) -- (cof.west);
        \node [above=0cm of cof, align=center] (cof_t) {Hypothetical CoF Table to explain \\ multiple image classifications};
    
    \end{tikzpicture}
    \caption{Component Diagram illustrating the process to produce a SCAP explanation, and how they are aggregated into CoF tables.}
    \label{tikz:comp_diagram}
\end{figure*}

\section{\large Creating Counterfactual Frequency Tables}


Our methodology for creating \textbf{Co}unterfactual \textbf{F}requency (CoF) tables is rooted in leveraging counterfactual explanations with semantic meaning. Initially, we introduce a framework aimed at generating \textbf{S}emantic \textbf{C}ounterfactuals for \textbf{A}ccurate \textbf{P}icture (SCAP) explanations. Building upon this, our key contribution is the presentation of CoF tables as a novel method for detecting learned shortcuts in models. These tables, by showcasing trends in model decisions through the aggregation of SCAP explanations, offer insight into the model's decision-making process and reveal potential shortcuts. This methodological approach not only identifies shortcuts but also aids in understanding the semantic underpinnings of model decisions.

\subsection{\textbf{S}emantic \textbf{C}ounterfactuals for \textbf{A}ccurate \textbf{P}icture (SCAP) Explanations}

Unlike other counterfactual explanations for images, SCAP explanations provide labels for the counterfactual-causing segments, which are essential for producing CoF tables. These labels must have semantic meaning to ensure meaningful aggregation based on the label. \\
\noindent The process to generate a SCAP explanation is outlined as follows (also illustrated in Figure~\ref{tikz:comp_diagram}):
\begin{enumerate}
    \item \textbf{Segmentation}: This step involves dividing the input image into multiple segments, each labelled according to its content. These segments are expected to carry semantic significance, which can be achieved through the application of pre-trained foundational segmentation models. 
    For many tasks, general purpose segmentation models, trained on datasets like COCO, will not provide the relevant segments, and thus not the aimed for counterfactuals.
    This is why the segmentation method can be anything that fulfils the requirements, hard coded rules, a single model or an ensemble of models.
    This also allows SCAP explanations to improve as segmentation models improve.
    The requirement is straightforward: an image is input, and a set of semantically meaningful segments with labels is output. In this work we demonstrate the use of multiple different segmentation methods.
    \item \textbf{Image Editing}: Following segmentation, this phase applies specific edits to each segment of the image. SEDC-T~\cite{vermeire2022explainable} opted to use only Gaussian blur, arguing that it caused minimal semantic change. However, the advent of pre-trained generative infill models has since altered this perspective. Given that models such as CNNs tend to bias towards texture over shape~\cite{geirhos2018imagenet}, we contend that blurring may not produce ideal counterfactuals for many segment types. For instance, while blurring snow or still water may have little effect, blurring a highly textured segment such as fur can result in significant changes. In this work, we demonstrate the efficacy of various editing techniques, including blurring, colour shifts, and generative infill. Edit functions should be decided in a problem-specific manner, as different edits provide different types of information. For example, blurring or infill can be used to ``remove'' a segment, while infill or colour shifts can be used to transform a segment into something entirely different.
    \item \textbf{Classification}: Any black box classifier. 
    From a practical perspective, we provide model wrappers for some common classifier frameworks: PyTorch, TensorFlow, and Scikit-learn, facilitating ease of integration within our codebase. 
    We additionally demonstrate the production of SCAP explanations using pre-trained models from Hugging Face.
    \item \textbf{Counterfactual Search}: Controls how the counterfactuals are found. This could be an optimisation procedure, looking to minimise some utility function, for example finding a minimal number of segments that when edited create a counterfactual. In this paper we only present simplistic counterfactuals of one segment, and report all counterfactuals we find.
\end{enumerate}

We opt for a broad interpretation of counterfactuals, reporting any modification that leads to a change in class.
This approach enables the generation of multiple counterfactuals per image, which can highlight instances where the classifiers predictions are particularly fragile. 

We chose to use counterfactuals for this method for a number of reasons:
\begin{itemize}
    \item They are always correct. The counterfactual instance we provide, is by definition always classified as a different class.
    \item As they are defined by a class change for a given model, they are always faithful.
    \item As we can create different counterfactuals by changing the edit function, we can reveal different aspects about the predictions. For example, does shifting the colour of a segment, or changing water segments to grass create counterfactuals?
\end{itemize}

\subsection{\textbf{Co}unterfactual \textbf{F}requency (CoF) Tables}

Let \(C\) denote the set of classes for which the model is trained to classify images, and let \(I = \{i_1, i_2, \ldots, i_n\}\) represent the dataset of images. For each image \(i_x \in I\), a segmentation method identifies a set of \(m_x\) segments \(S_x = \{S_x^1, S_x^2, \ldots, S_x^{m_x}\}\), each tagged with a semantic label. The CoF table aggregates and quantifies the frequency with which segments, identified by their labels, lead to a change in the model's predicted class when edited, indicating a counterfactual instance.

The construction of a CoF table for a set of images \(I\) aggregates the frequency of how often each segment label produces a counterfactual across \(I\).
Specifically, for a segment labelled \(l\) that appears in at least one image within \(I\), we define its Counterfactual Frequency as:

\begin{displaymath}
CoF(l) = \sum_{x=1}^{n} \sum_{j=1}^{m_x} \mathbf{1}(label(S_x^j) = l \land g(i_x, S_x^j) \neq c_x)
\end{displaymath}

where:
\begin{itemize}
    \item \(label(S_x^j) = l\) verifies that the \(j\)-th segment in image \(i_x\) is labeled \(l\),
    \item \(g(i_x, S_x^j) \neq c_x\) checks whether editing segment \(S_x^j\) results in a classification different from the original predicted class \(c_x\) for image \(i_x\),
    \item \(\mathbf{1}(\cdot)\) is the indicator function, equal to 1 when its argument is true (the relevant edit caused a class change), and 0 otherwise.
\end{itemize}.

A CoF table is defined as a comprehensive aggregation of \(CoF(l)\) values for every unique segment label \(l\) identified across the entire dataset \(I\). Formally, the CoF table can be represented as:

\begin{displaymath}
\text{CoF Table} = \left\{ (l, CoF(l)) \mid l \in L \right\}
\end{displaymath}

where \(L\) is the set of all unique segment labels identified by the segmentation model across the dataset \(I\).

Computationally, our implementation is calculated per image (or image batch) rather than per label.
For each image within the dataset, our method attempts to find SCAP explanations, any that are found, are recorded in a result table, which is the precursor to CoF tables.
Optionally, additional information can be added to each row, such as the initial and counterfactual classification, the position of the counterfactual segment in the image, and the ground truth label. 
This information can then be used to tailor CoF tables to provide specific insights.

For instance, one might generate a CoF table exclusively detailing misclassifications or specifically those counterfactuals that rectify a misclassification. Revisiting the watermark example, consider if each watermarked image displays the watermark in any of the four corners, rather than one position. Positional information regarding the watermark could then be leveraged to determine whether the model exhibits a preference for shortcuts associated with certain locations, or if it is equally influenced across various positions.

In this work we show how large set pre-trained segmentation models can enable SCAP to detect shortcuts through CoF tables. The COCO dataset has 133 possible labels for panoptic segmentation, in this work we show that models trained from this are sufficient to detect shortcuts in some circumstances. However, this limited pool of labels is lacking for some problem types, for example, models trained on the COCO panoptic set cannot detect a watermark. 
This can be combated by using task specific segmentation models, or even some open set segmentation models with prompts. 

This leads us to our proposed use of CoF tables, as a test result. 
The test must be well-designed to yield useful results. If the goal is to test for background or object shortcuts within the closed-set, a general model with blurring may suffice. 
However, if the aim is to test for reliance on segments outside of this set, or those with textures resistant to blurring, a different segmentation or editing method must be employed. 
Furthermore, the choice of edit function must be logically sound; for instance, merely blurring a watermark might not adequately obscure it, whereas applying generative infill to the affected area is likely to produce a more faithful image.

Observing the similarities between our analysis and the domain of visual analytics, we advocate for a methodology that adheres to Schneiderman's visual information-seeking mantra: `Overview first, zoom and filter, then details on demand' \cite{Shneiderman2003Eyes}. 
The initial presentation of the CoF table serves as the overview, where overarching trends can be identified which, in a well crafted test, can uncover shortcuts.
The `zoom and filter' stage involves examining subsets of the CoF tables—such as those focusing on misclassifications, corrected classifications, or specific positions to enhance understanding of reliance on shortcuts. Finally details on demand can be looking at the actual counterfactual instances themselves.

We posit that this approach offers a more refined and effective means to identify shortcuts and provides insights into strategies for their mitigation. Leveraging the findings from SCAP explanations, the same  segmentation models and edit functions could be employed to augment training data for a more robust classifier.










%% file: Sections/4-results.tex
\section{\large Results}



In this section we demonstrate the usage of CoF tables to identify shortcuts in a number of datasets. For each dataset, we show individual SCAP explanations and discuss the findings from their respective CoF tables.
To illustrate the flexibility of our methodology we utilise
a variety of classifiers, segmentation methods, and edit
functions. 



\subsection*{Colourised MNIST}

We begin with a straightforward example to illustrate background bias using the MNIST dataset. 
To achieve this, we converted the original single-channel greyscale images to three-channel RGB format. 
Each class in the dataset was assigned a unique, visually distinct colour. 
For each image, pixels that were completely black (RGB values of (0,0,0)) were changed to the colour assigned to their class. 
This modification creates an extremely simple shortcut for models to learn, providing a clear starting point for our analysis.

\begin{figure}[h]
    \includegraphics[width=\linewidth]{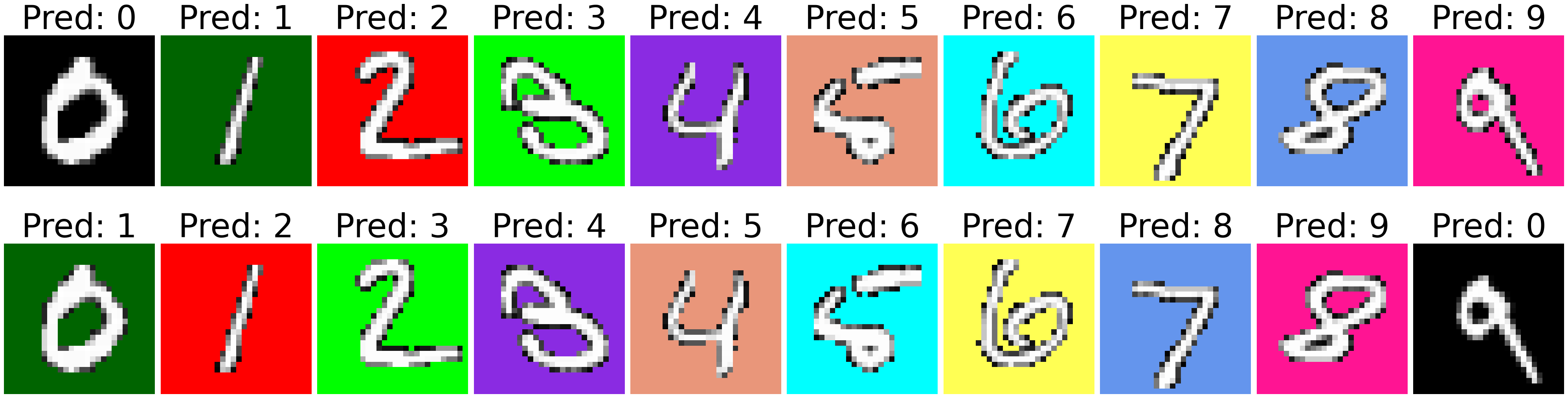}
    \caption{In the first row, example images from our biased MNIST dataset, with the their classifications.
             In the second row, their counterfactual counterparts, with the new classification.}
    \label{fig:cMNIST}
\end{figure}

\begin{figure}[h]
    \centering
    \begin{subfigure}[t]{0.32\linewidth}
        \centering
        \includegraphics[width=\linewidth]{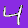}
        \caption{\centering Predicted class: 4}
    \end{subfigure}
    \hfill 
    \begin{subfigure}[t]{0.32\linewidth}
        \centering
        \includegraphics[width=\linewidth]{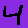}
        \caption{\centering Counterfactual Segment: Background}
    \end{subfigure}
    \hfill 
    \begin{subfigure}[t]{0.32\linewidth}
        \centering
        \includegraphics[width=\linewidth]{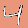}
        \caption{\centering Counterfactual instance: 5}
    \end{subfigure}
    \caption{Individual SCAP explanation from the biased MNIST dataset.}
\end{figure}

\definecolor{color0}{RGB}{0,0,0}
\definecolor{color1}{RGB}{0,100,0}
\definecolor{color2}{RGB}{255,0,0}
\definecolor{color3}{RGB}{0,255,0}
\definecolor{color4}{RGB}{138,43,226}
\definecolor{color5}{RGB}{233,150,122}
\definecolor{color6}{RGB}{0,255,255}
\definecolor{color7}{RGB}{255,255,84}
\definecolor{color8}{RGB}{100,149,237}
\definecolor{color9}{RGB}{255,20,147}

\begin{table}[h]
    \centering
    \caption{Multi-Class CoF table for the coloured MNIST dataset.
    Segmenting the background by colour and editing the image by shifting the colour.
    It demonstrates a consistent shortcut of background colour for predicting labels
    of the trained model.}
    \begin{tabular}{lll}
        \hline
        \textbf{Class} & \textbf{Segment Name} & \textbf{Frequency} \\
        \hline
        0 & \tikz\draw[color0,fill=color0] (0,0) circle (.5ex); background & 100\% \\
        1 & \tikz\draw[color1,fill=color1] (0,0) circle (.5ex); background & 100\% \\
        2 & \tikz\draw[color2,fill=color2] (0,0) circle (.5ex); background & 100\% \\
        3 & \tikz\draw[color3,fill=color3] (0,0) circle (.5ex); background & 100\% \\
        4 & \tikz\draw[color4,fill=color4] (0,0) circle (.5ex); background & 100\% \\
        5 & \tikz\draw[color5,fill=color5] (0,0) circle (.5ex); background & 100\% \\
        6 & \tikz\draw[color6,fill=color6] (0,0) circle (.5ex); background & 100\% \\
        7 & \tikz\draw[color7,fill=color7] (0,0) circle (.5ex); background & 100\% \\
        8 & \tikz\draw[color8,fill=color8] (0,0) circle (.5ex); background & 100\% \\
        9 & \tikz\draw[color9,fill=color9] (0,0) circle (.5ex); background & 100\% \\
        \hline
    \end{tabular}
\end{table}

Using this biased dataset, we trained a simple Convolutional Neural Network (CNN).
After only a few epochs, the model achieves near 100\% accuracy in both the training and testing set, thanks to the simple shortcut. 

We do not use a model as our segmentation method for this experiment, but a hard-coded function. We identify the background by determining the most common pixel value in the image, and then finding all pixel indexes that have that colour, providing us with a single segment per image. Using the most common pixel value, we are also able to estimate the class with 100\% accuracy. The edit function then simply changes the colour of all pixels in the segment provided. We chose to shift the colour to the colour of the next class, which produces the distinct pattern seen in figure~\ref{fig:cMNIST}.

This toy dataset provides us with a baseline to clearly outline the aims of CoF tables, showing a clear, ever-present shortcut without any noise. Additionally, it highlights a previously discussed major flaw in saliency maps, which is the lack of information on why certain pixels are important. Individual saliency maps would likely show that the background was indeed being used as a shortcut. However, they wouldn't be able to identify the colour bias, which is key to understanding the shortcut.

\subsection*{Biased Action Recognition (BAR)}
\label{sec:BAR}

The Biased Action Recognition (BAR) dataset~\cite{nam2020learning} comprises a multi-class classification with six classes. Each class is defined by a specific action (e.g., climbing, diving), with the training dataset exhibiting a bias towards particular environments—climbing actions all occur on rock walls, while diving is primarily represented in underwater settings. 
This environmental bias is exclusive to the training dataset. In contrast, the testing dataset introduces a shift in bias; for instance, images in the climbing class transition from outdoor rock walls to indoor, artificial climbing walls.

\begin{figure}[h]
    \centering
    \begin{subfigure}[t]{0.32\linewidth}
        \centering
        \includegraphics[width=\linewidth]{images/climb_vis_a.jpg}
        \caption{\centering Predicted class: Rock Climbing}
    \end{subfigure}
    \hfill 
    \begin{subfigure}[t]{0.32\linewidth}
        \centering
        \includegraphics[width=\linewidth]{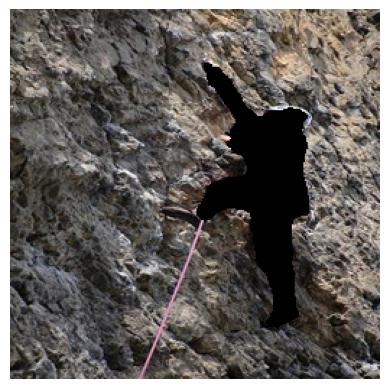}
        \caption{\centering Counterfactual Segment: Rock}
    \end{subfigure}
    \hfill 
    \begin{subfigure}[t]{0.32\linewidth}
        \centering
        \includegraphics[width=\linewidth]{images/climb_vis_d.jpg}
        \caption{\centering Counterfactual instance: Throwing}
    \end{subfigure}
    \caption{Individual SCAP explanation from the BAR dataset.}
    \label{fig:BAR_ind}
\end{figure}

\begin{table}[h]
    \centering
    \caption{CoF table for the climbing class of the BAR dataset.
    Shows all segments that cause at least 4\% of counterfactuals.}
    \begin{tabular}{ll}
    \hline
    \textbf{Segment Name} & \textbf{Frequency} \\ \hline
    rock          & 46.2\%          \\
    tree          & 11.7\%          \\
    dirt         & 6.7\%            \\
    mountain     & 6.3\%            \\
    unrecognised & 5.4\%            \\
    stone wall   & 4.0\%             \\ \hline
    \end{tabular}
    \label{tbl:BAR_Top3_Climb}
\end{table}


We trained a simple TensorFlow CNN on this dataset, achieving 99.43\% accuracy on the training set but only 21.41\% on the testing set.
This significant difference in accuracy highlights the model's reliance on shortcuts derived from the background features within the training data, rather than learning generalisable features.

We then search for SCAP explanations for all images in the training dataset, using our method.
For segmentation, we employed the pre-trained DETR panoptic segmentation model\cite{carion2020end}.
Any sections of the image without an assigned label by DETR are labelled afterwards as `unrecognised', ensuring each image is fully represented through the segments derived from it.
As our edit function, we applied a simple Gaussian blur across the entire counterfactual segment, as illustrated in Figure \ref{fig:BAR_ind}.

Table~\ref*{tbl:BAR_Top3_Climb} illustrates our method for identifying the reliance on rock backgrounds as a shortcut within the climbing class, utilising CoF tables. We identified 21 segment names responsible for generating counterfactuals and consequently narrowed down the CoF table to only those segments that constitute at least 4\% of the counterfactuals identified. Notably, over 46\% of all counterfactuals stem from rock backgrounds, indicating a clear reliance on the shortcut, without the need for the shifted testing data.


\subsection*{ImageNet Boats}

\begin{figure}[h]
    \centering
    \begin{subfigure}[t]{0.32\linewidth}
        \centering
        \includegraphics[width=\linewidth]{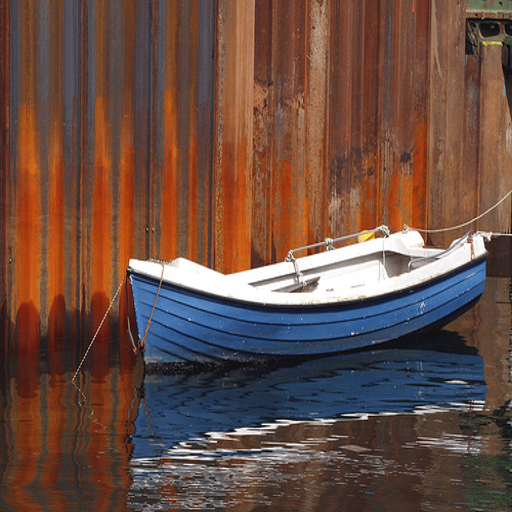}
        \caption{\centering Original Image}
    \end{subfigure}
    \hfill 
    \begin{subfigure}[t]{0.32\linewidth}
        \centering
        \includegraphics[width=\linewidth]{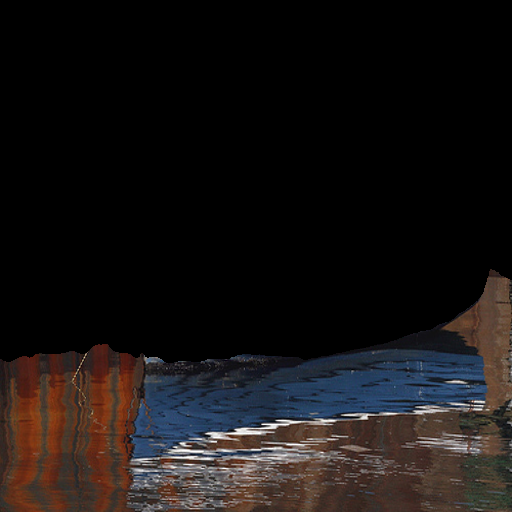}
        \caption{\centering Counterfactual Segment: Water}
    \end{subfigure}
    \hfill 
    \begin{subfigure}[t]{0.32\linewidth}
        \centering
        \includegraphics[width=\linewidth]{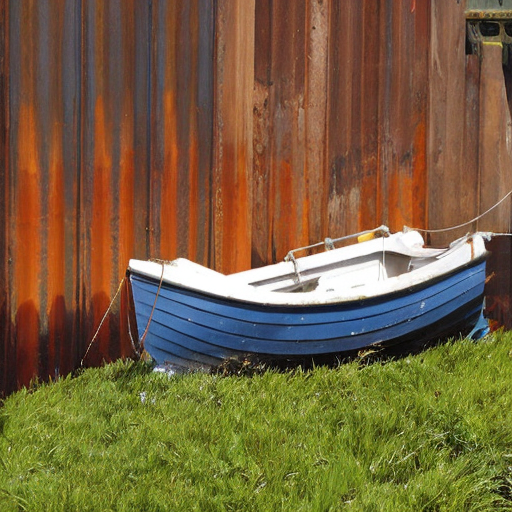}
        \caption{\centering Counterfactual instance: Tub}
    \end{subfigure}
    \caption{Individual SCAP explanation for rowing boat class of ImageNet.}
\end{figure}

\begin{table}[h]
    \centering
    \caption{CoF table showing the percentage of river vs sea counterfactual segments for the boat classification}
    \begin{tabular}{ll}
    \hline
    \textbf{Segment Name} & \textbf{Frequency} \\ \hline
    river          & 55.1\%            \\
    sea          & 44.9\%         \\ \hline
    \end{tabular}
\end{table}

\begin{table}[h]
    \centering
    \caption{CoF table showing the percentage of which a counterfactual is created when replacing a given segment within an image with grass.}
    \begin{tabular}{ll}
    \hline
    \textbf{Segment Name} & \textbf{Frequency} \\ \hline
    river          & 68.8\%            \\
    sea          & 71.5\%         \\ \hline
    \end{tabular}
\end{table}

In this experiment, we explore the `rowing boat' category from the ImageNet dataset~\cite{deng2009imagenet}. We utilise the well-studied ResNet-50 model~\cite{he2016deep}, pre-trained on the ImageNet-1k, as our classifier.
Similar to the approach taken in the BAR experiment, we employ the DETR segmentation model to provide segments. 
However, in this instance, we only consider the label synonymous with water.
In this example, those are `river' and `sea', due to the absence of a generic `water' label widely recognised in these images. 
Instead of blurring, we use generative inpainting as our edit function. 
To accomplish this inpainting, we use a pre-trained version of Stable Diffusion 2 for inpainting~\cite{Rombach_2022_CVPR}.

To effectively expose the underlying background bias within our dataset, we create counterfactual images by replacing water segments with infill prompted by `grass'.

Grass or fields are also likely to be highly common in different classes, so we decide to replace a background strongly associated with boats, water, with one that is common for other classes.
 Considering the frequent association of boats with water and the prevalence of grass or open fields in various other classes, this methodological choice serves as a deliberate contrast. By substituting the background closely associated with the boat category—water—with one commonly found in numerous other categories—grass—we aim to highlight the model's reliance on background cues for classification.

\subsection*{Imagenet Horses and Zebras}

\begin{figure}[h]
    \centering
    \begin{subfigure}[t]{0.32\linewidth}
        \centering
        \includegraphics[width=\linewidth]{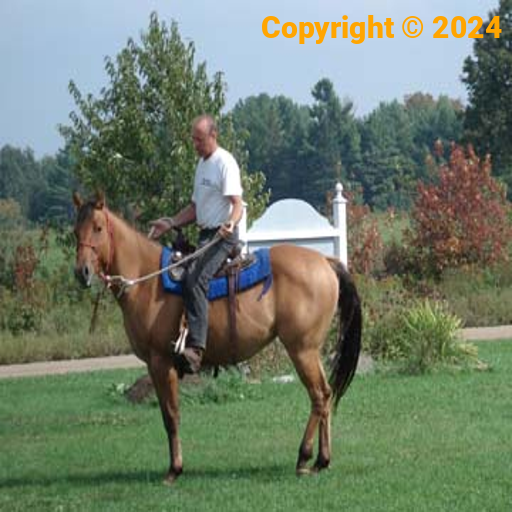}
        \caption{\centering Predicted class: Horse}
    \end{subfigure}
    \hfill 
    \begin{subfigure}[t]{0.32\linewidth}
        \centering
        \includegraphics[width=\linewidth]{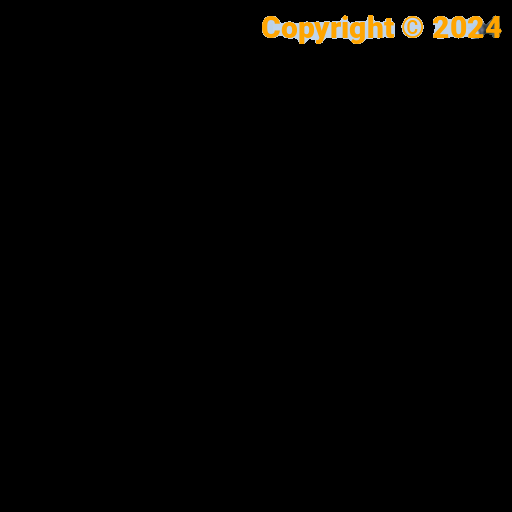}
        \caption{\centering Counterfactual Segment: Watermark}
    \end{subfigure}
    \hfill 
    \begin{subfigure}[t]{0.32\linewidth}
        \centering
        \includegraphics[width=\linewidth]{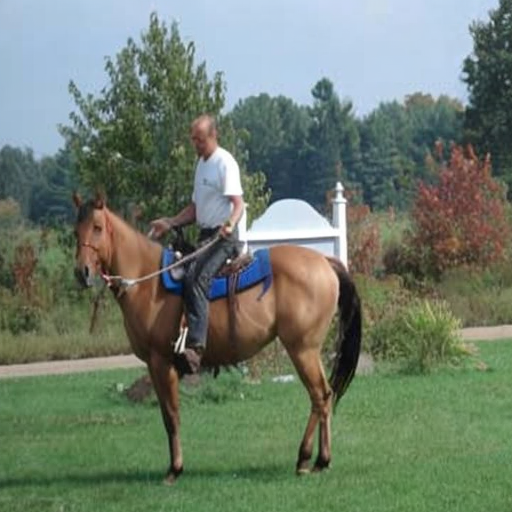}
        \caption{\centering Counterfactual instance: Zebra}
    \end{subfigure}
    \caption{Individual SCAP explanation for Horse vs Zebra classification.}
\end{figure}

\begin{table}[h]
\centering
\caption{CoF table showing the percentage of images in the training set for which we find counterfactuals for specific segments. Only showing segments that cause counterfactuals in over 1\% of images.}
\begin{tabular}{ll}
\hline
\textbf{Segment Name} & \textbf{Frequency} \\ \hline
tree                  & 5.3\%              \\
grass                 & 4.6\%              \\
horse                 & 2.9\%              \\
person                & 2.3\%              \\
sky                   & 2.1\%              \\
fence                 & 2.0\%              \\
watermark             & 1.2\%              \\ \hline
\end{tabular}
\label{tbl:horse_segment_frequency}
\end{table}

\begin{table}[h]
\centering
\caption{CoF table showing the frequency that when a given segment is detected, it causes a counterfactual. Only showing frequencies of above 15\%. Limited to only segments that occur in at least 20 images.}
\begin{tabular}{ll}
\hline
\textbf{Segment Name} & \textbf{Frequency} \\ \hline
watermark             & 24.5\%             \\
banner                & 23.8\%             \\
rock                  & 21.7\%             \\
sand                  & 21.2\%             \\
tree                  & 16.3\%             \\
snow                  & 16.0\%             \\
mountain              & 15.2\%             \\ \hline
\end{tabular}
\label{tbl:horse_segment_importance}
\end{table}

\begin{table}[h]
    \centering
    \caption{CoF table with position of watermark segments included.}
    \begin{tabular}{lll}
    \hline
    \textbf{Segment Name} & \textbf{Position} & \textbf{Frequency} \\ \hline
    watermark         & bottom left & 29.6\%            \\
    watermark         & top left & 25.9\%            \\
    watermark         & bottom right & 22.2\%            \\
    watermark         & top right & 22.2\%            \\
    \hline
    \end{tabular}
    \label{tbl:horse_watermark_pos}
\end{table}


In our subsequent experiment, we return to the ImageNet dataset, selecting the `mare, female horse' and `zebra' classes for analysis. 
This choice is informed by our intention to simulate a controlled environment that mirrors the `watermark shortcut' observed in the PASCAL VOC 2007 dataset, where watermarks are leveraged as classification shortcuts for the horse category, with 15-20\% of images containing watermarks~\cite{lapuschkin2019unmasking}. To replicate this scenario, we introduce watermarks into 10\% of the training images within the horse class.

To segment the watermarks, we use Language Segment-Anything, a model built on top of Segment-Anything~\cite{kirillov2023segment}. Segment-Anything is an open-set segmentation model, meaning it can segment a huge number of different objects. However, it doesn't assign semantic labels for these segments.
Language Segment-Anything addresses this limitation by not only providing semantic labels for the identified segments but also supporting user-defined prompts. For our analysis, we specify the prompt `watermark' for each image, enabling precise identification and labelling of watermark segments.
Additionally to this, we use the DETR segmentation model to provide general segments.

For the editing process, we once again employ generative inpainting as our edit function, for watermarks this time without providing any prompt, thereby allowing the model to autonomously determine the most fitting infill for each image. For the segments provided by DETR we blur the segments, as there is too much variance in size and importance of segments to reliably infill without a prompt.

Following this, we train a model for binary classification between the two chosen classes, achieving an accuracy of around 95\% in both training and testing accuracy.

In table~\ref{tbl:horse_watermark_pos}, we show a CoF table with the position of the watermarks. For this experiment, watermarks were present in 25\% of the images, with an even distribution across each of the four corners. Although the difference is very small, these results suggest a slight skew towards the left-hand side of the image, motivating further analysis.

Table~\ref{tbl:horse_segment_frequency} presents the standard CoF table, showing the frequency with which certain segments trigger counterfactuals within the training dataset.
However, given that many segments rarely appear in images, this result can be misleading.
Common segments such as grass, horse (DETR often misclassifies zebras as horses), and tree, appear in most images, naturally leading to a greater number of counterfactuals associated with them. 
To address this, table~\ref{tbl:horse_segment_importance} refines the analysis by showing the frequency at which the detection of a segment results in a counterfactual.

This underscores the importance of the watermark shortcut. Watermarks are present in only 10\% of images from the horse class, and consequently, in just 5\% of the total dataset. This scarcity makes detection challenging for many techniques, which often struggle to identify features that are infrequently represented in the training data.

The removal of watermarks results in counterfactuals in approximately 1\% of images, indicating that manual detection would require extensive analysis. Furthermore, methods designed to aggregate numerous explanations are likely to identify several other factors before considering watermarking.

The use of CoF tables facilitates the exploration of such explanations, enabling the discovery of critical insights like these.

%% file: Sections/5-end.tex
\section{\large Discussion}

\begin{figure}
    \centering
    \includegraphics[width=0.8\linewidth]{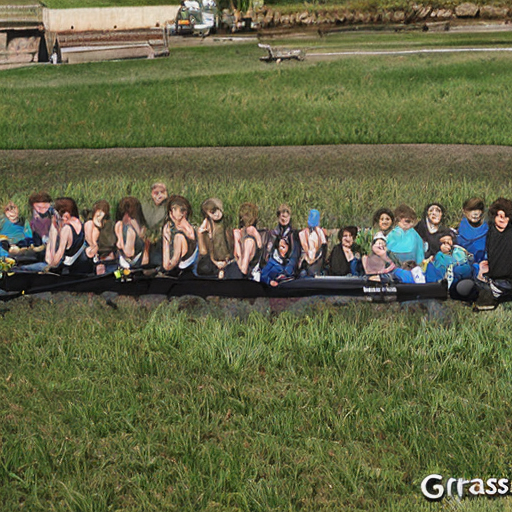}
    \caption{Image showing the result of inpainting the water in an image with the prompt grass. Unwantedly, the inpainting model has edited the faces of the people on the boat, and added text on top of the grass.}
    \label{fig:boat_failure}
\end{figure}

However, our method has a number of shortcomings.
Firstly, the quality of SCAP explanations is directly tied to the suitability of the segmentation method and edit function used. Although open-set segmentation models and generative infill methods show promise with adaptable fidelity and an ability to create counterfactuals for a wide array of problem types, they introduce variability.
The open-set segmentation model we use provides incorrect segments more frequently than its closed set counterparts. 
The generative inpainting we use can create corrupted images such as those seen in figure~\ref{fig:boat_failure}, and it is difficult to automatically detect this.
However, we notice that these distortions are minimal when the segment is small, and there is no prompt.
We show how these issues can be mitigated through thoughtful test design, and we are hopeful that as both segmentation and image editing techniques improve, the capacity for SCAP explanations will too.

As with many counterfactual explanation methods, SCAP suffers from both issues of coverage and the disagreement problem. We attempt to embrace the latter, suggesting that if an image can provide many counterfactuals, it is a particularly fragile image and warrants further investigation.

Given the capacity of CoF tables to unveil shortcuts with an overview, we advocate for their inclusion in model cards. Although the tests utilised would need verification, standardised procedures could be employed to identify known shortcuts in widely used datasets.

In summary, this study presents a novel method to identify shortcuts in machine learning models. Our contributions can be summarised as follows:
\begin{itemize}
    \item The outlining of Semantic Counterfactuals for Accurate Picture (SCAP) explanations as a way of incorporating semantic meaning to counterfactual explanations.
    \item The introduction of Counterfactual Frequency (CoF) tables, a way of aggregating instance-based SCAP explanations to find trends across groups of data.
    \item Demonstrating the capability of identifying learned shortcuts across a number of datasets using CoF tables, including multiple examples that current techniques would find difficult to detect.
\end{itemize}

\section*{Acknowledgement}
This work was supported by the AXA Joint Research Initiative (JRI) project ``Explainable Artificial Intelligence'' 
and received funding from the Flemish Government under the ``Onderzoeksprogramma Artificiële Intelligentie (AI) 
Vlaanderen''.


%% file: Sections/appendix.tex
\clearpage
\section*{Appendix}

\begin{table}[h]
    \centering
    \caption{Top 3 most common segment names leading to counterfactuals for other classes excluding climbing in the BAR dataset.}
    \begin{tabular}{lll}
    \hline
    \textbf{Class}        & \textbf{Segment Name} & \textbf{Count} \\ \hline
    Diving                & water-other           & 2              \\
                          & teddy bear            & 1              \\
                          & grass          & 1              \\ \hline
    Fishing               & sea                   & 6              \\
                          & river                 & 5              \\
                          & tree           & 4              \\ \hline
    Pole Vaulting         & person                & 7              \\
                          & sky-other      & 5              \\
                          & tree           & 3              \\ \hline
    Racing                & car                   & 49             \\
                          & motorcycle            & 24             \\
                          & truck                 & 18             \\ \hline
    Throwing              & person                & 85             \\ \hline
    \end{tabular}
    \label{tbl:BAR_other_top3}
\end{table}

\begin{figure}[hb]
    \includegraphics[width=\linewidth]{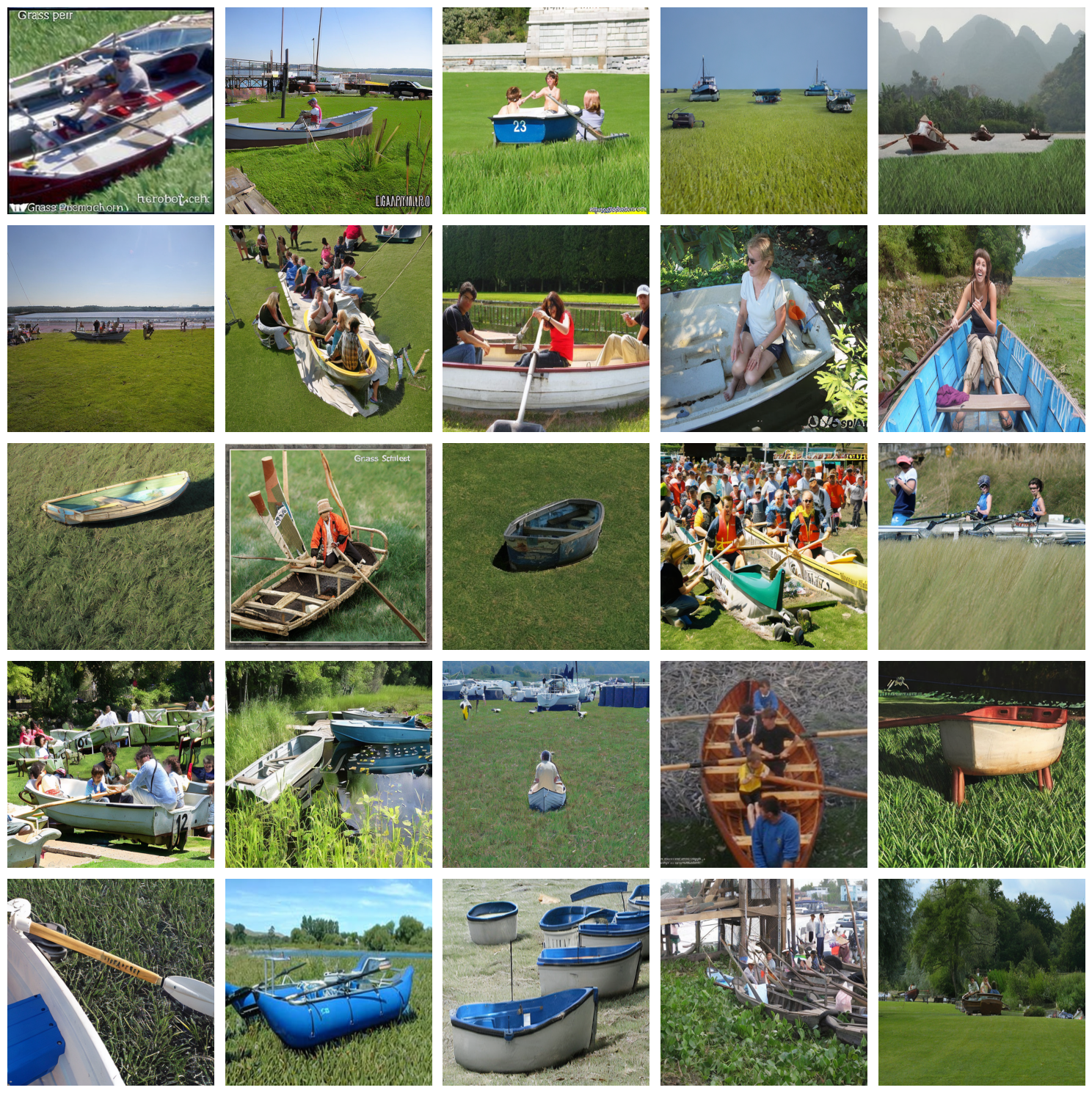}
    \caption{Examples of counterfactual images for ResNet-50, generated by infilling grass to segments labeled as 
    `river' and `sea' by the DETR segmentation model within images from the `paddle boat' category of ImageNet. 
    }
    \label{fig:boatMatrix}
\end{figure}

\begin{figure}[hb]
	\includegraphics[width=\linewidth]{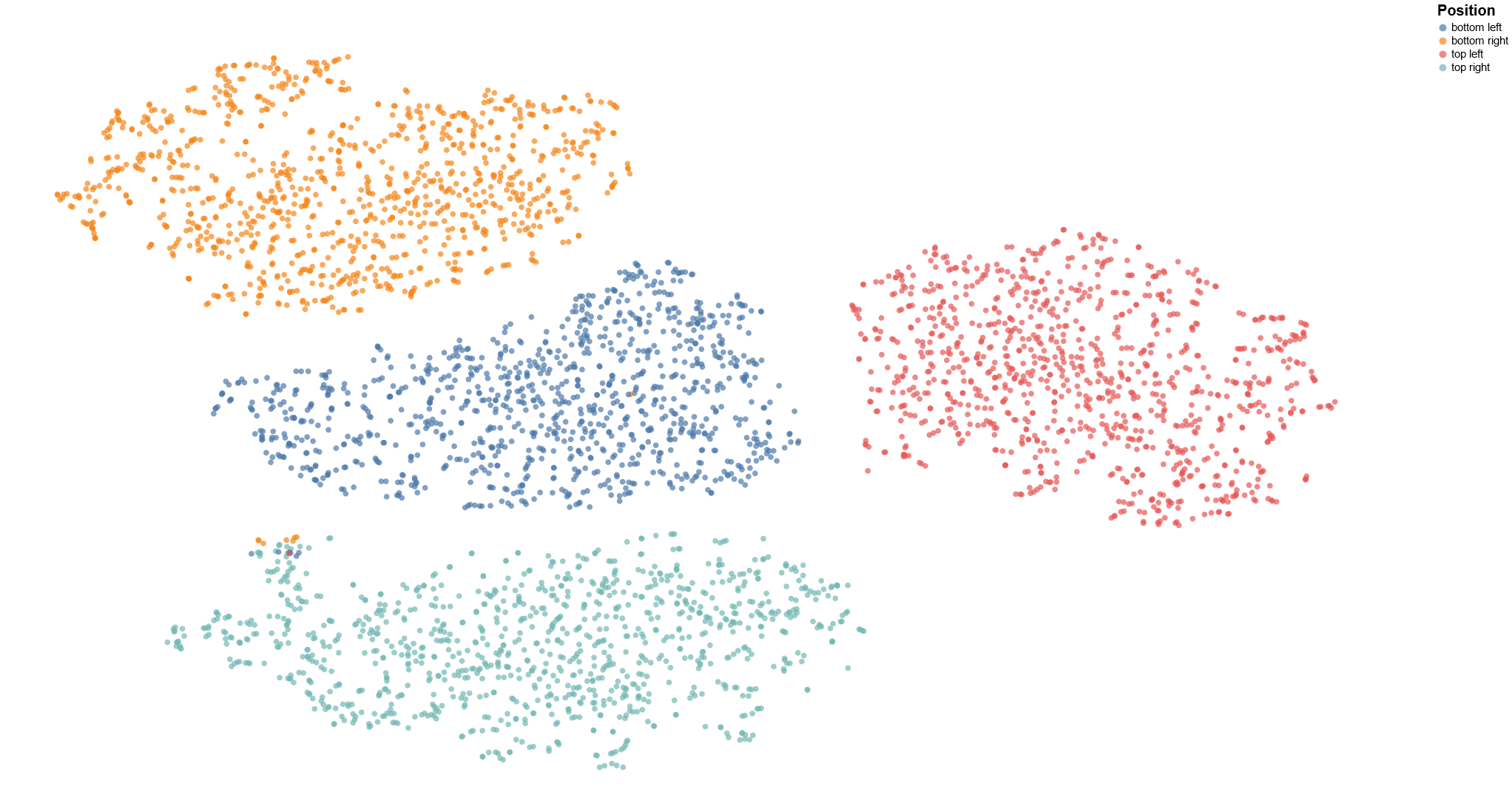}
	\caption{This figure presents a t-SNE scatter plot illustrating the spectral clustering results 
	from the SpRAy method applied to a subset of MNIST images at 4 distinct positions. 
	Each point in the plot corresponds to an individual image, colour-coded to reflect its 
	position (as one of the four corners).
	This visualisation highlights the predominance of position over the shape of the image in spectral clustering.}
	\label{fig:SprayVis}
\end{figure}